\documentclass[runningheads]{llncs}
\usepackage{graphicx}
\usepackage{array}
\usepackage{collcell}
\usepackage{colortbl}
\usepackage{amsmath,amssymb} 
\usepackage{color}

\newcolumntype{P}[1]{>{\centering\arraybackslash}p{#1}}

\usepackage{subfig}

\begin{document}
\title{Fully-Convolutional Point Networks\\for Large-Scale Point Clouds} 

\titlerunning{Fully-Convolutional Point Networks\\for Large-Scale Point Clouds}
%
\author{Dario Rethage\inst{1} \and
Johanna Wald\inst{1} \and
J\"urgen Sturm\inst{2} \and \break
Nassir Navab\inst{1} \and
Federico Tombari\inst{1}}
%
\authorrunning{D. Rethage, J. Wald, J. Sturm, N. Navab and F. Tombari}
%

\institute{Technical University of Munich \\
\email{dario.rethage@tum.de, johanna.wald@tum.de, navab@cs.tum.de, tombari@in.tum.de}\\
\and
Google \\
\email{jsturm@google.com}}
\maketitle              
\begin{abstract}
This work proposes a general-purpose, fully-convolutional network architecture for efficiently processing large-scale 3D data. One striking characteristic of our approach is its ability to process unorganized 3D representations such as point clouds as input, then transforming them internally to ordered structures to be processed via 3D convolutions. In contrast to conventional approaches that maintain either unorganized or organized representations, from input to output, our approach has the advantage of operating on memory efficient input data representations while at the same time exploiting the natural structure of convolutional operations to avoid the redundant computing and storing of spatial information in the network. The network eliminates the need to pre- or post process the raw sensor data. This, together with the fully-convolutional nature of the network, makes it an end-to-end method able to process point clouds of huge spaces or even entire rooms with up to $200k$ points at once. Another advantage is that our network can produce either an ordered output or map predictions directly onto the input cloud, thus making it suitable as a general-purpose point cloud descriptor applicable to many 3D tasks. We demonstrate our network's ability to effectively learn both low-level features as well as complex compositional relationships by evaluating it on benchmark datasets for semantic voxel segmentation, semantic part segmentation and 3D scene captioning.

\end{abstract}

\section{Introduction}

\begin{figure}[t]
    \centering
 \includegraphics[width=\linewidth]{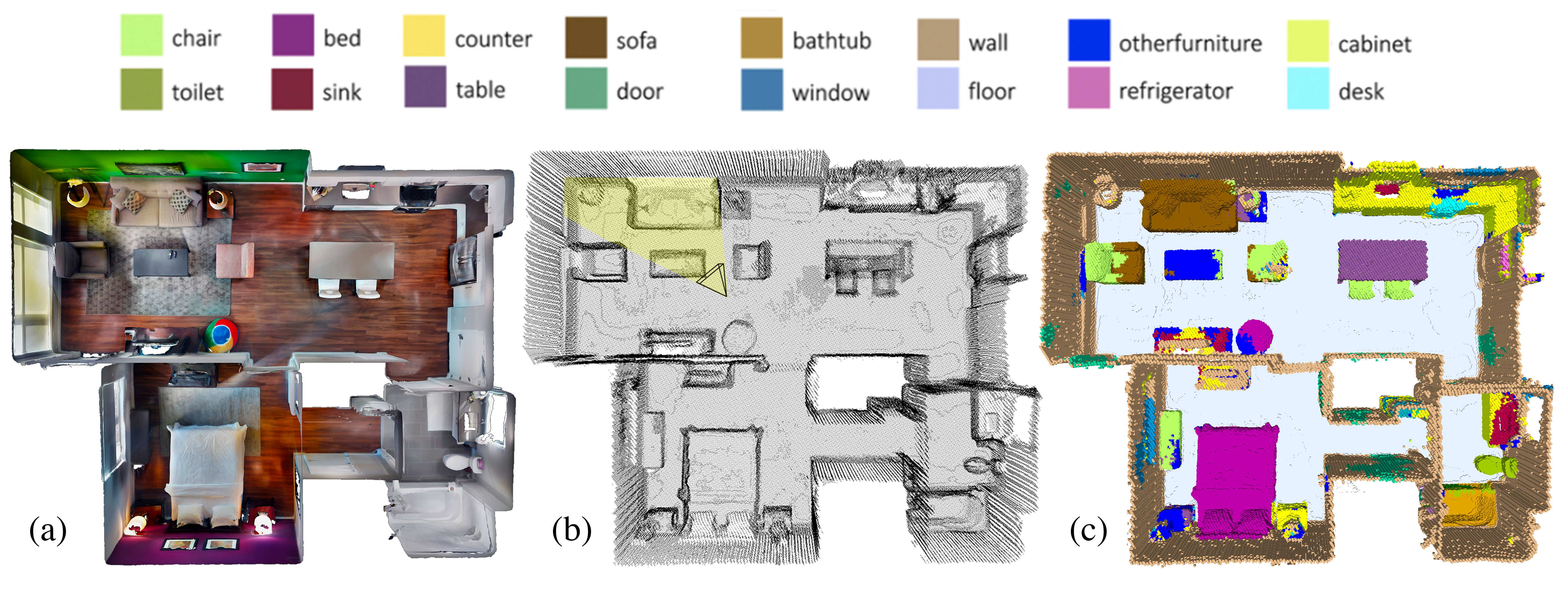} 
    \caption{Example result of our FCPN on semantic voxel labeling and captioning on an Tango 3D reconstruction / point cloud: (a) 3D reconstruction (not used), (b) Input point cloud (c) Output semantic voxel prediction. A possible caption for the camera pose in (b) is \textit{``There is a place to sit in front of you"}}
    \label{fig:teaser} 
\end{figure}

Processing 3D data as obtained from 3D scanners or depth cameras is fundamental to a wealth of applications in the field of 3D computer vision, scene understanding, augmented/mixed reality, robotics and autonomous driving. 
The extraction of reliable semantic information from a 3D scene is useful, for example, to appropriately add virtual content to the 3D space around us or describe it to a visually-impaired person. 
Analogously, in robotics, processing of 3D data acquired from a depth camera allows the robot to perform sophisticated tasks, beyond path-planning and collision avoidance, that require intelligent interaction in real world environments.

A recent research trend has focused on designing effective learning architectures for processing common 3D data representations such as point clouds, meshes and voxel maps, to be employed in tasks such as voxel-based semantic scene segmentation \cite{Dai2017}, part-based segmentation of 3D objects \cite{Qi2017} and 3D correspondence matching \cite{Zeng2016}. A primary objective of these models is robustness against typical issues present when working with real world data such as, noise, holes, occlusion and partial scans, as well as viewpoint changes and 3D transformations (rotation and translation). Another challenge more related to semantic inference relates to dealing with the large number of classes that characterize real world scenarios and their typically large intra-class variance. 


In pursuit of a versatile 3D architecture, applicable in small- and large-scale tasks, it is not only necessary to extract meaningful features from 3D data at several scales, but also desirable to operate on a large spatial region at once. For this, fully-convolutional networks (FCN) \cite{Shelhamer2016} have recently grown to prominence due to their drastic reduction in parameters and flexibility to variable input sizes. However, learning these hierarchical statistical distributions starting at the lowest level requires a huge amount of data. To achieve this, some methods train on synthetic data, but suffer from the domain gap when applied to the real-world \cite{Song2016}. A big step toward closing this gap is ScanNet, a large-scale dataset of indoor scans \cite{Dai2017}. Methods that achieve state-of-the-art performance on these challenging tasks quickly reach the memory limits of current GPUs due to the additional dimension present in 3D data \cite{Klokov2017,Riegler2017}. While the aforementioned FCN architecture reduces the number of parameters, it requires the input to be in an ordered (dense) form. To bypass the need to convert the raw, unordered data into an ordered representation, PointNet \cite{Qi2017} proposes an architecture that directly operates on sets of unordered points. Since PointNet only learns a global descriptor of the point cloud, Qi et. al later introduced a hierarchical point-based network with PointNet++ \cite{Qi2017_2}. While achieving impressive results in several tasks, PointNet++ cannot take advantage of the memory and performance benefits that 3D convolutions offer due to its fully point-based nature. This requires PointNet++ to redundantly compute and store the context of every point even when they spatially overlap.\\

\begin{figure}[h]%
    \centering
    \subfloat[1.5m, 8k points]{{\includegraphics[height=3.5cm]{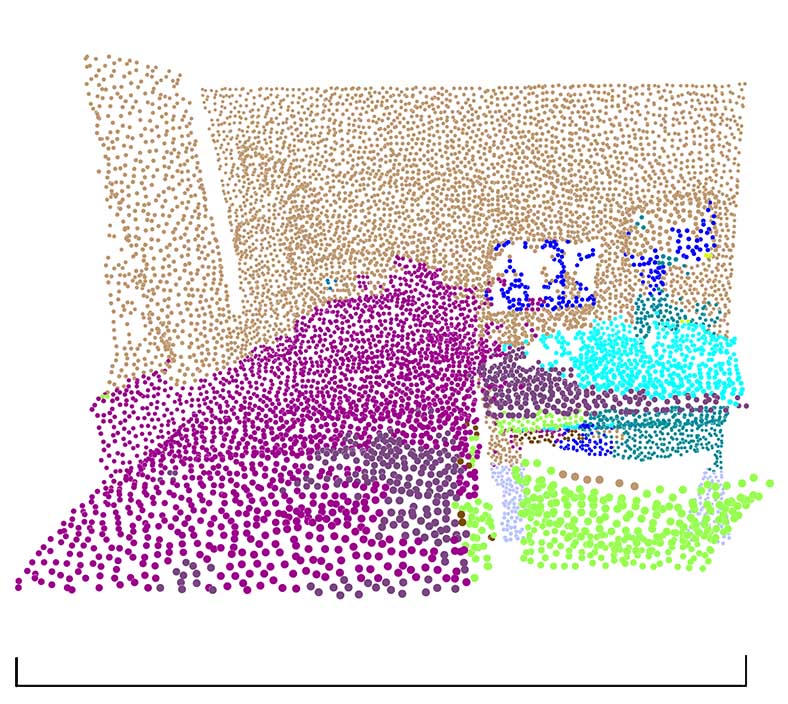}}
    \label{fig:scale_results_a}}%
    \subfloat[2.4m, 16k points]{{\includegraphics[height=3.5cm]{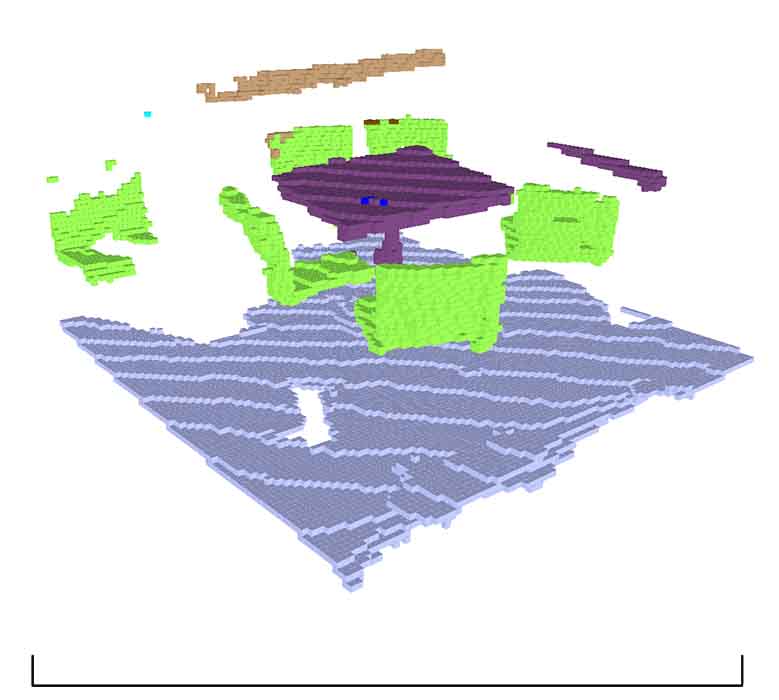}}\label{fig:scale_results_b}}%
    \subfloat[10m, 200k points]{{\includegraphics[height=3.5cm]{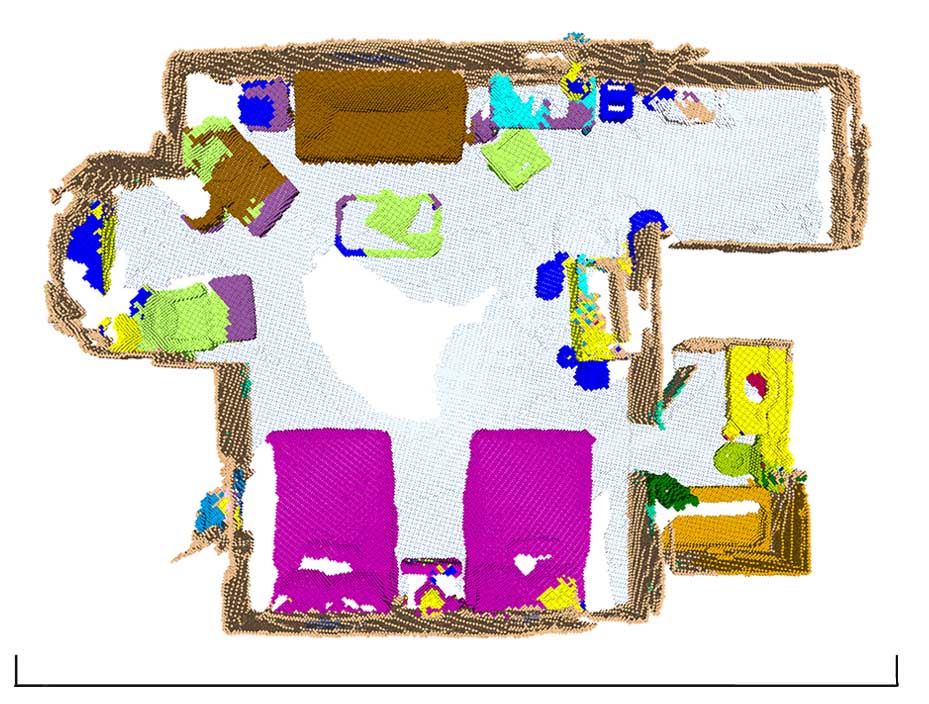}}\label{fig:scale_results_c}}%
    \caption{Visualization of the semantic segmentation on (a) a depth image, (b) a $2.4m \times 2.4m \times 2.4m$ partial reconstruction (c) an entire reconstruction of a hotel suite. Please note that each of these outputs are predicted in a single shot from the same network trained with $2.4m \times 2.4m \times 2.4m$ volumes}%
    \label{fig:scale_results}%
\end{figure}

We present a general-purpose, fully-convolutional network architecture for processing 3D data: Fully-Convolutional Point Network (FCPN). Our network is hybrid, i.e. designed to take as input unorganized 3D representations such as point clouds while processing them internally in an organized fashion through 3D convolutions. This is different from other approaches, which require both input and internal data representation to be either unorganized point sets \cite{Qi2017,Qi2017_2,Manessi2018} or organized data \cite{Song2016,Dai2017}. The advantage of our hybrid approach is to take the benefits of both representations. Unlike \cite{Song2016,Dai2017}, our network operates on memory efficient input representations that scale well with the scene/object size and transforms it to organized internal representations that can be processed via convolutions. 
A benefit of our method is that it can scale to large volumes while processing point clouds in a single pass. It can also be trained on small regions, e.g. $2.4 \times 2.4 \times 2.4$ meters and later applied to larger point clouds during inference. A visualization of the output at three different scales of our network trained on semantic voxel labeling is given in Figure \ref{fig:scale_results}.
While the proposed method is primarily intended for large-scale, real-world scene understanding applications, demonstrated by the semantic voxel labeling and 3D scene captioning tasks, the method is also evaluated on semantic part segmentation to demonstrate its versatility as a generic feature descriptor capable of operating on a range of spatial scales.

Our main contributions are (1) a network based on a hybrid unorganized input/organized internal representation; and, (2) the first fully-convolutional network operating on raw point sets. We demonstrate its scalability by running it on full ScanNet scenes regions of up to $80m^2$ in a single pass. In addition, we show the versatility of our learned feature descriptor by evaluating it on both large-scale semantic voxel labeling as well as on 3D part segmentation. In addition, as a third contribution, we explore the suitability of our approach to a novel task which we dub \textit{3D captioning}, that addresses the extraction of meaningful text descriptions out of partial 3D reconstructions of indoor spaces.
We demonstrate how our approach is well suited to deal with this task leveraging its ability to embed contextual information, producing a spatially ordered output descriptor from the unordered input, necessary for captioning. For this task, we also publish a novel dataset with human-annotated captions.


\section{Related Work}

Deep learning has already had a substantial impact on 3D computer vision. Those 3D deep learning approaches can be categorized as either (a) volumetric or voxel-based models that operate on an ordered input (see Section \ref{section:voxel-networks}) or (b) point-based models that work entirely with unordered data (see Section \ref{section:point-networks}).
Some approaches do not deal directly with 3D data, but instead operate in 2D or 2.5D (RGB-D), for example, multi-view CNNs \cite{Su2015,Qi2016,Kaiming2017}. Typically, RGB(-D) based methods put more emphasis on color information and less on geometry. This makes it less robust under varying lighting conditions. Instead, our proposed approach is fully geometric and we therefore do not further review RGB-D based methods here.

\subsection{Voxel-based networks}
\label{section:voxel-networks}

Most volumetric or voxel-based 3D deep learning methods have proven their value by achieving state of the art accuracy \cite{Dai2017,Maturana2015,Cicek2016,Milletari2016,Song2016} on a variety of tasks. These methods employ convolutional architectures to efficiently process data. This however requires the input data to be organized -- stored in a dense grid of predefined order. Each uniformly-sized voxel in this grid is then labeled with a semantic class. Ordered representations of 3D data have the advantage of featuring constant time (O(1)) lookup of neighbors. Such a representation usually explicitly models empty space making it memory-intensive. This is particularly inefficient since 3D data is often very sparse. Further, voxelization imposes an explicit resolution on the data. To transform sparse 3D data to a dense representation requires preprocessing: either using a simple occupancy grid or by encoding it, for example, in a truncated signed-distance field (TSDF) as done in KinectFusion\cite{Izadi2011,Song2016}. This means the model does not see the data itself, but a down-sampled encoding of it.\\

VoxNet was a pioneering effort using 3D convolutional networks for object recognition \cite{Maturana2015}.
Similarly, Wu et. al learn deep volumetric representations of shapes for shape recognition and completion \cite{Wu2015}. Another popular example that applied to medical imaging is 3D U-Net \cite{Cicek2016}. It processes at a relatively high input resolution of $132 \times 132 \times 116$, but only outputs the center of the volume. With current memory availability, voxelization of larger spaces generally requires labeling at a lower sampling density implying a loss of resolution. Alternatively, a higher density can be achieved if a smaller context is used to inform the prediction of each voxel. For example, ScanNet \cite{Dai2017} performs semantic voxel labeling of each approximately $5cm$ voxel column in a scene using the occupancy characteristics of the voxel's neighborhood.
SSCNet achieves a larger spatial extent of $2.26m^3$ for jointly semantically labeling and completing depth images by use of dilated convolutions. However, also in this scenario, the size as well as resolution of the output is reduced \cite{Song2016}.
\\

To address this limitation, OctNet propose to use OctTrees, known to efficiently partition 3D space in octants, in a deep learning context by introducing convolutions directly on the OctTree data structure \cite{Riegler2017}. Klokov and Lempitsky demonstrate another solution using kd-trees \cite{Klokov2017}. However, these methods still impose a minimum spatial discretization. Our network is flexible and efficient enough to mitigate this memory limitation without discretizing the input in any way.

\subsection{Point-based networks}
\label{section:point-networks}

A pioneering work that operates on unordered points directly is PointNet \cite{Qi2017}. Qi et. al showed the advantages of working with point sets directly, learning more accurate distributions in continuous space, thereby bypassing the need to impose a sampling resolution on the input. PointNet achieves state-of-the-art performance on classification, part segmentation and scene segmentation tasks while operating at a single-scale. They further claim robustness to variable point density and outliers. However, since PointNet does not capture local features the authors later introduced PointNet++ \cite{Qi2017_2} 
a multi-scale point-based network that uses PointNet as a local feature learner for semantic segmentation, among other tasks. In the semantic segmentation task, point contexts are first abstracted and later propagated through 3NN interpolation in the latent space. For the larger-scale scene segmentation task, it is relevant that PointNet++ redundantly processes and stores the context of points throughout the network. This prevents PointNet++ from being able to process a large space in a single pass. Instead in the semantic scene segmentation task, it processes a sliding region of $1.5 \times 1.5 \times 3$ meters represented by 8192 points.\\ 

Our work is -- as the first of its kind -- hybrid and therefore positioned between these volumetric and point-based methods.
As such, allowing powerful multi-scale feature encoding by using 3D convolutions while still processing point sets directly.

\section{Fully-Convolutional Point Network}

\begin{figure}[h]
    \centering
 \includegraphics[width=\linewidth]{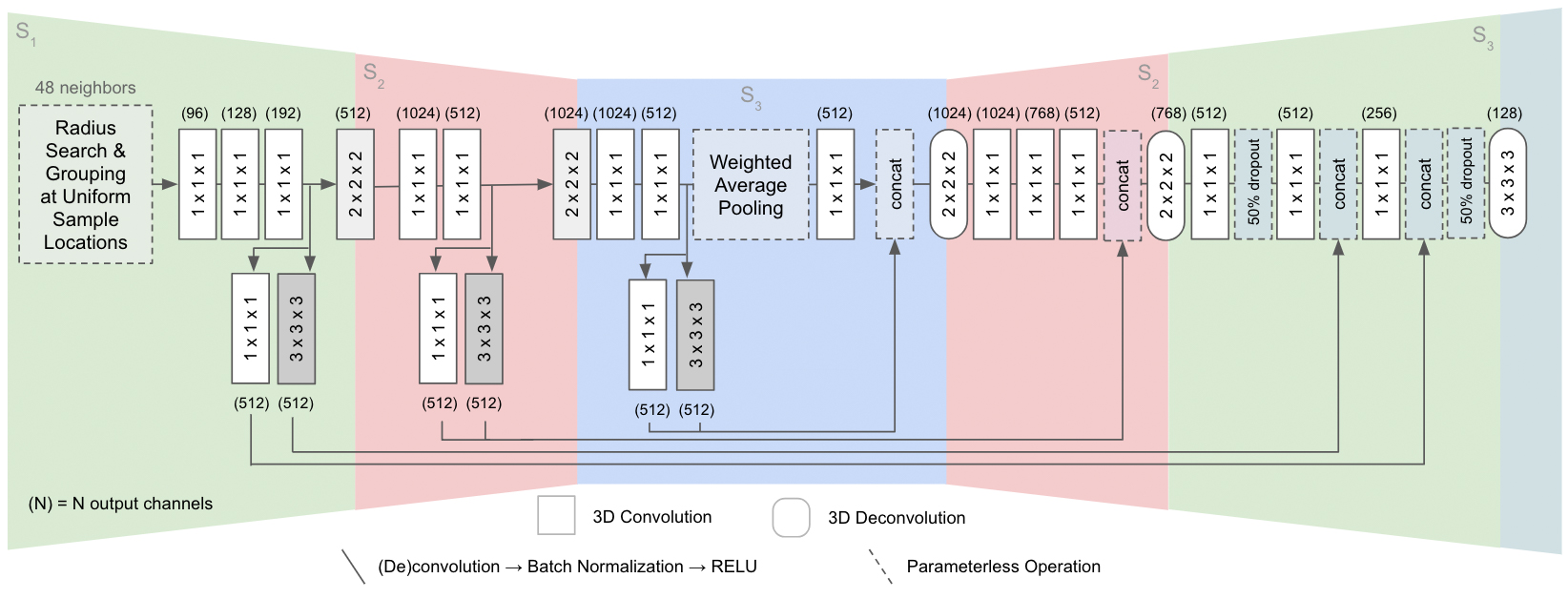} 
    \caption{Fully-Convolutional Point Network Architecture}
    \label{fig:architecture} 
\end{figure}



\noindent
Our network operates on unorganized input and employs PointNet \cite{Qi2017} as a low-level feature descriptor. Contrary to PointNet++ \cite{Qi2017_2}, a uniform sampling strategy is applied. This step captures the precise local geometry in each local region and transforms the unordered input internally to an ordered representation for further processing. This transformation is followed by 3D convolutions to then learn compositional relationships at multiple scales. Our network abstracts the space at three scales $S_1, S_2$ and $S_3$. Skip connections with $1\times 1\times 1$ and $3\times 3\times 3$ convolutions at each scale inexpensively double the total number of feature scales the network captures. At the highest abstraction scale, the features are additionally average pooled together weighted by their distance to each voxel. Once features are produced at each scale of abstraction, feature volumes of the same size are concatenated and progressively upsampled by 3D deconvolutions to achieve the desired output sampling density. Figure \ref{fig:architecture} gives an overview of the proposed architecture. Depending on the scenario, additional convolutions, latent nearest-neighbor interpolation or fully-connected layers can be applied to produce an ordered, end-to-end point mapping or single-value output, respectively. In the following sections, the different components of our network are described in more detail.\\

\subsection{Architecture}

The Fully-Convolutional Point Network consists of four main modules: a series of abstraction layers, feature learners at different scales, a weighted-average pooling layer, and a merging stage where responses are hierarchically merged back together.

\subsubsection{Abstraction Layers}
Three abstraction layers are used to achieve a hierarchical partitioning, both spatially and conceptually. The first level captures basic geometric features like edges and corners, the second level responds to complex structure and the highest level to structure in context of other structures.\\


The first level employs a simplified PointNet \cite{Qi2017_2}, proven to efficiently capture the geometry in a local region. It consists of a radius search \& grouping, $1\times1\times1$ convolutions followed by a max-pooling layer. Applying PointNet in a uniformly spaced 3D grid produces a 3D feature volume representing the lowest level physical features. This feature volume feeds into the next abstraction layer. Higher level abstraction layers are implemented as 3D convolutions with a kernel size and stride of 2. They are designed to abstract the space in a non-overlapping fashion with 8 features (octants) of the preceding abstraction layer being represented by a single cell in the subsequent layer, just like an OctTree. This non-overlapping spatial-partitioning significantly reduces the memory required to store the space at each abstraction level.

\subsubsection{Feature Learners}

\begin{figure}[t]
    \centering
 \includegraphics[width=\linewidth]{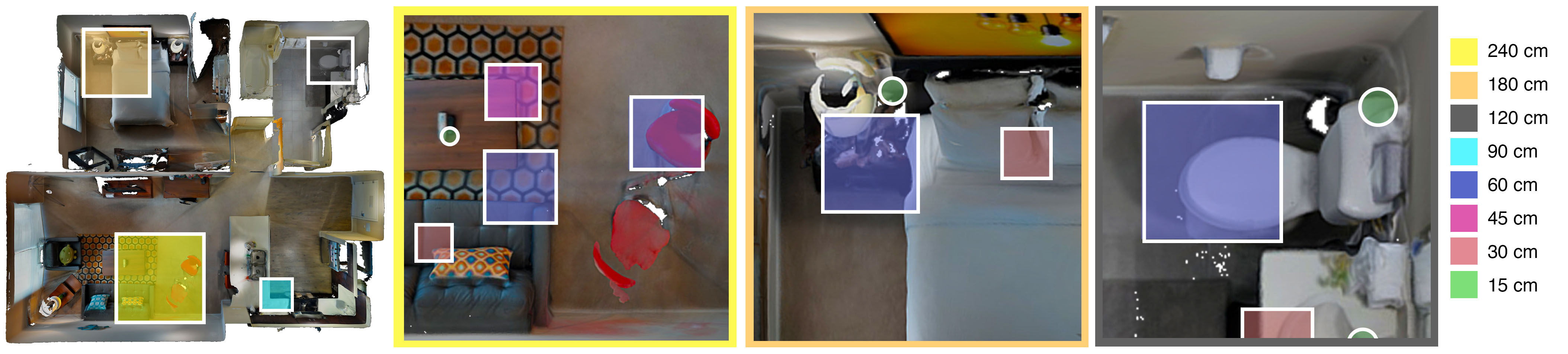}
    \caption{Visualization of the different spatial scales used to encode indoor spaces within our model on an example apartment scan. For simplification we only show their 2D top-down view. Further, please note that 15cm features (green) are spherical while the others are cubic}
    \label{fig:multiscale}  
\end{figure}

With three levels of abstraction, we now employ $1 \times 1 \times 1$ and $3 \times 3 \times 3$ convolutions to extract meaningful features at more scales (see Figure \ref{fig:multiscale}). For each abstraction level, skip connections propagate features at the level's inherent spatial scale as well as $3 \times$ it to be merged later in the network. This allows the network to better recognize structures at a wider range of scales and to overcome the strictly non-overlapping partitioning of the abstraction layers. 

\subsubsection{Weighted Average Pooling}

The weighted average pooling layer cost-effectively incorporates long-range contextual information. For every cell in the highest abstraction level, the responses of all other cells in the space are averaged together weighted by their euclidean distance to a 1m sphere around the cell. Thus, cells positioned closest to the surface of this sphere are weighted most heavily. This puts emphasis on long-range contextual information, instead of information about directly adjacent cells which is already captured by the respective $3\times 3\times 3$ skip connection. This improves the discriminative capability of the network by allowing neighboring semantics to influence predictions. For example, distinguishing chairs and toilets by considering the presence of a desk or rather a sink nearby. The parameterless nature of this layer is not only extremely cost-effective, but provides a more informative signal. The average spatial pooling effectively removes the exact configuration of the structures in the vicinity, while retaining their semantic identities. This is a desirable characteristic because the identity of nearby objects or structures help discriminate boundary cases more so than the configuration they are in. In the example of the chair/toilet, it is informative to know the presence of a sink nearby much more than the fact that the sink is to the right of the toilet. We also avoid an inherent challenge: larger-scale spatial contexts ($>$1m) encourage a model to learn entire configurations of spaces, which does not lead to strong generalizability. Finally, the average weighted pooling layer exhibits the flexibility required to scale the network up to larger spaces during inference.

\subsubsection{Merging}
In the merging stage, skip connections corresponding to each abstraction level are first concatenated and then upsampled to $2\times$ their spatial resolution by 3D deconvolutions. This allows the features at each abstraction level to be progressively merged into each other. $1\times 1\times 1$ convolutions add expressive power to the model between deconvolutions.

\subsection{Output Representations}

Several variants of the network are suitable for different scenarios: For producing organized output, the network is fitted with an additional deconvolution layer to produce the desired output point density. Latent nearest-neighbor interpolation is used in applications where semantics are mapped, for end-to-end processing, to each point in the input cloud. Fully-connected layers are appropriate when summarizing the entire input such as in the case of scene captioning.\\

\subsection{Uniform vs. Furthest Point Sampling}

Furthest point sampling is very effective for describing the characteristics of structure (occupied space) because it makes no prior assumptions as to the spatial distribution of the data. However, for describing entire spaces (occupied + unoccupied), a uniform sampling is the only way to ensure we consider every part of it.



\subsection{Full-Volume Prediction}

Fully-Convolutional Point Network labels the full spatial volume it is given as input. It achieves this by upsampling feature maps to the original sampling density as well as symmetrically padding feature volumes before performing $3\times 3\times 3$ convolutions. This is validated by the fact that regions directly outside of the input volume are most likely to exhibit the same occupancy characteristics as the closest cell within it, since occupancy characteristics present at the edges of the volume are likely to extend beyond it.

\subsection{Scalability}

The network is flexible in that it can be trained on smaller training samples, then be scaled up during inference to process spaces multiple times larger than it was trained on in a single shot. The network successfully predicts a ~$80$m$^2$ space consisting of ~$200k$ points at once. An even larger spatial extent can be processed at sparser point density. This further extends the versatility of the network for other use cases such as autonomous driving.

\section{3D Captioning}
\label{section:Captioning}

We introduce a new scene understanding task we call 3D Captioning: generating meaningful textual descriptions of spaces. We envision this as being useful for assistive technology, in particular for the visually impaired when navigating and interacting in unfamiliar environments. To test the model's proficiency on this task, we create a dataset of human-annotated captions based on ScanNet \cite{Dai2017}, we select the top 25 sentences best describing the diversity of spaces found in ScanNet. They are designed to answer 3 types of questions: "\textit{What is the functional value of the space I'm in?}", "\textit{How can I move?}", and "\textit{How can I interact with the space?}". Every 100th frame of a scene is annotated with 0 or more applicable captions. The dataset was then validated to remove outliers. To accomplish this, a Scene Caption Annotation Tool was built and used to annotate roughly half of ScanNet. We release this dataset together with the source code. The statistics of the captioning dataset are given in the supplementary material. 

\section{Evaluation}





We evaluate our method on (a) small-scale 3D part segmentation as well as (b) large-scale semantic segmentation tasks (see Section \ref{section:EvalSVL}). We evaluate semantic segmentation on ScanNet, a 3D dataset containing 1513 RGB-D scans of indoor environments with corresponding surface reconstructions and semantic segmentations \cite{Dai2017}. This allows us to compare our results against ScanNet's voxel-based prediction \cite{Dai2017} and PointNet++'s point cloud-based semantic segmentation \cite{Qi2017_2}. We achieve comparable performance, while -- due to our fully-convolutional architecture -- being able to process considerably larger spaces at once. Our second evaluation (b) in Section \ref{section:EvalPartSegmentation} on a benchmark dataset for model-based part segmentation shows our networks capability to generalize to other tasks and smaller scales. To further show the usefulness of a spatially ordered descriptor in higher-level scene understanding tasks, we train our network to predict captions for unseen scenes (see Section \ref{section:Captioning}) -- results for the 3D captioning task are presented here \ref{section:EvalCaptioniong}. 

\subsection{Semantic Voxel Labeling}
\label{section:EvalSVL}

In the semantic voxel labeling task, the network is trained to predict the semantics of the occupied space in the input from a set of 20 semantic classes. We present a variant of the network for Semantic Voxel Labeling along with the experimental setup. 

\subsubsection{Data Preparation}

Training samples are generated following the same procedure in ScanNet. We extract volumes exhibiting at least 2\% occupancy and 70\% valid annotations from 1201 scenes according to the published ScanNet train/set split. Training samples are $2.4m^3$ and feature a uniform point spacing of $5cm^3$. This produces a training set of $~75k$ volumes. During training, samples are resampled to a fixed cardinality of 16k points. Augmentation is performed on the fly: random rotation augmentation along the up-down axis, jitter augmentation in the range +/- 2cm and point dropout between 0-80\%. Only X,Y,Z coordinates of points are present in the inputs. Ground-truths consist of 20 object classes and 1 class to represent unoccupied space. Each scene in the 312 scene test set is processed by predicting $2.4m^3$ cutouts of the scene. Each semantic class is weighted by the inverse log of its per-point frequency in the dataset.

\subsubsection{Network}

The three spatial scales S1, S2, S3 of the Semantic Voxel Labeling network are 15cm, 30cm, 60cm, respectively. As a result, the network extracts features at 15cm, 30cm, 45cm, 60cm, 90cm and 180cm scales and pools features together at the 60cm spatial scale. Three $1\times 1\times 1$ layers follow each abstraction, feature learning and upsampling layer. An additional deconvolution layer achieves a final output density of $5cm^3$. Dropout (50\%) is applied before this layer. We also employ a final $3\times 3\times 3$ convolution in the last layer to enforce spatial continuity in adjacent predictions, avoiding single point misclassification. The network is trained with an ADAM optimizer starting at a learning rate of 0.01, decaying by half every epoch, for 5 epochs.

\subsubsection{Results}

\begin{figure}[h!]
    \centering
 \includegraphics[width=0.9\linewidth]{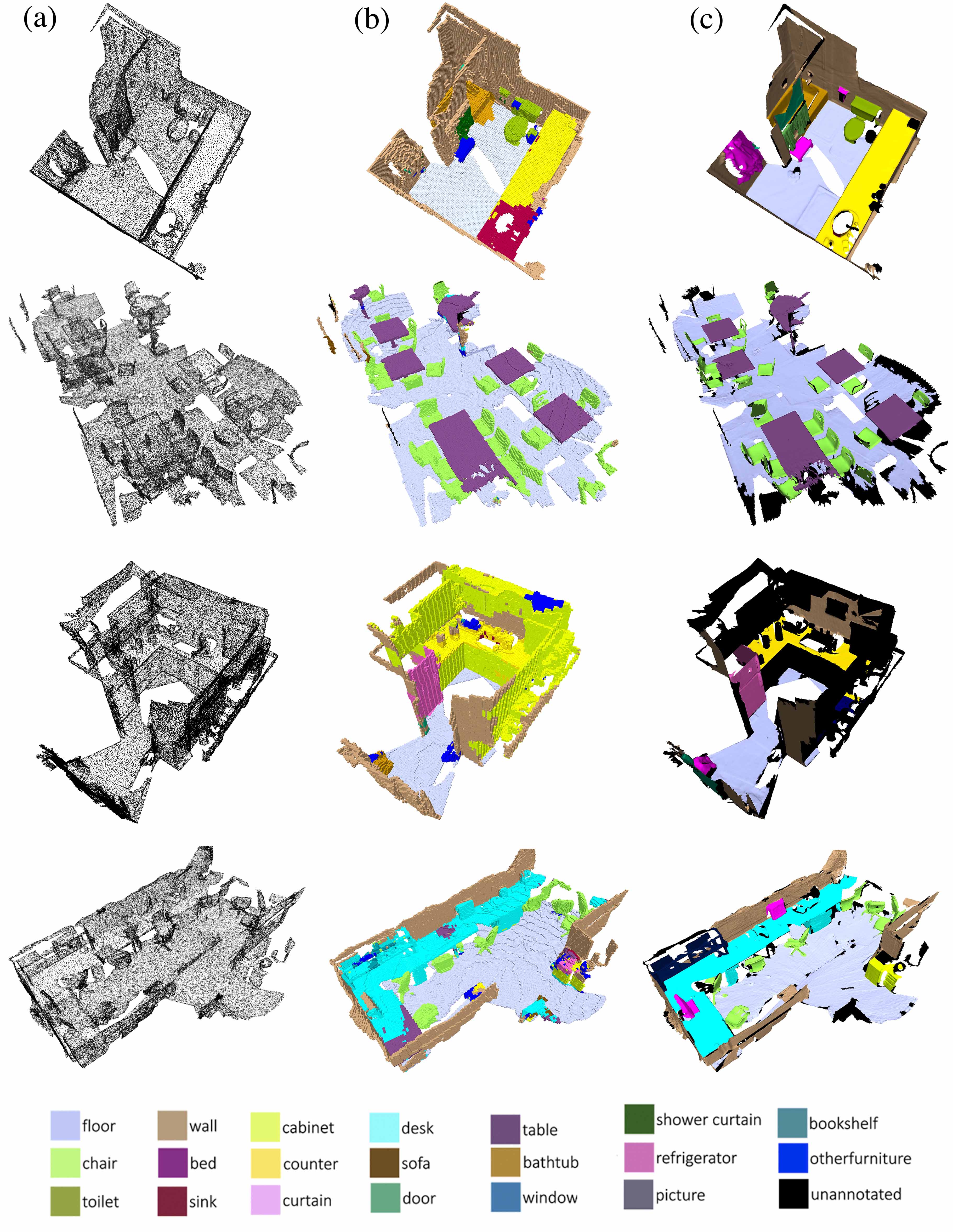}
    \caption{Qualitative results of the semantic voxel labeling on an example scene of ScanNets test sequences. (a) Input Point Cloud, (b) Semantic Voxel Prediction by our FCPN and (c) Ground Truth Semantic Annotation}
    \label{fig:qualitative_results} 
\end{figure}

Table \ref{table:scannet_eval} gives quantitative results of our semantic segmentation on a voxel-basis for 20 classes. Our method achieves a weighted accuracy of 82.6\% and an unweighted accuracy of 54.2\%; in comparison ScanNet only labels $73\%$ (weighted) or $50.8\%$ (unweighted) of the voxels correctly. The best performing of three PointNet++ variants (MSG+DP) reports $84.5\%$ (weighted) or $60.2\%$ (unweighted). Our method outperforms ScanNet by a large margin particularly in the classes: desk, toilets, chairs and bookshelves. Please note, that our method has the advantage of being able to process all scenes, from small bathrooms up to whole apartments, in a single shot in comparison to PointNet++ who combine the predictions of a sliding volume with a majority voting. Some qualitative results with corresponding ground truth annotations are shown in Figure \ref{fig:qualitative_results}.

\newpage

\definecolor{FloorColor}{RGB}{189, 198, 255}
\definecolor{WallColor}{RGB}{190, 153, 112}
\definecolor{ChairColor}{RGB}{152, 255, 82}
\definecolor{SofaColor}{RGB}{113, 77, 0}
\definecolor{TableColor}{RGB}{122, 71, 130}
\definecolor{DoorColor}{RGB}{0, 174, 126}
\definecolor{CabinetColor}{RGB}{213, 255, 0}
\definecolor{BedColor}{RGB}{158, 0, 142}
\definecolor{DeskColor}{RGB}{1, 255, 254}
\definecolor{ToiletColor}{RGB}{133, 169, 0}
\definecolor{SinkColor}{RGB}{149, 0, 58}
\definecolor{WindowColor}{RGB}{0, 125, 181}
\definecolor{PictureColor}{RGB}{107, 104, 130}
\definecolor{BookshelfColor}{RGB}{0, 143, 156}
\definecolor{CurtainColor}{RGB}{255, 166, 254}
\definecolor{ShowerCurtainColor}{RGB}{0, 100, 1}
\definecolor{CounterColor}{RGB}{255, 229, 2}
\definecolor{RefrigeratorColor}{RGB}{232, 94, 190}
\definecolor{BathtubColor}{RGB}{187, 163, 0}
\definecolor{OtherFurnitureColor}{RGB}{0, 0, 255}

\begin{table}[h]
\centering
\caption{Semantic voxel label prediction accuracy on ScanNet test scenes. Please note, ScanComplete only trains on 6 of the 20 classes present in the ScanNet test set}
\label{table:scannet_eval}
\resizebox{\textwidth}{!}{%
{\renewcommand{\arraystretch}{1.3}%
\begin{tabular}{p{0.5cm}p{0.1cm}p{3.5cm}p{3cm}p{2.3cm}p{2.8cm}p{2.5cm}p{2.3cm}}
 & & Class & \% of Test Scenes & ScanNet\cite{Dai2017} & ScanComplete\cite{dai2018scancomplete} & PointNet++\cite{Qi2017_2} & FCPN (Ours)\\
\hline
\cellcolor{FloorColor} && Floor	&	35.7\%	&	90.3\% &90.2\% & \textbf{97.8\%} & 96.3\% \\\arrayrulecolor{white}\hline
\cellcolor{WallColor} && Wall	&	38.8\%	&	70.1\% &88.8\%	& \textbf{89.5\%} & 87.7\% \\\arrayrulecolor{white}\hline
\cellcolor{ChairColor} && Chair	&	3.8\%	&	69.3\% & 60.3\% & \textbf{86.0\%} & 81.6\% \\\arrayrulecolor{white}\hline
\cellcolor{SofaColor} && Sofa	&	2.5\%	&	75.7\% &72.5\%	& 68.3\% & \textbf{76.0\%} \\\arrayrulecolor{white}\hline
\cellcolor{TableColor} && Table	&	3.3\%	&	\textbf{68.4\%} &n/a	& 59.6\% & 67.6\% 
\\\arrayrulecolor{white}\hline
\cellcolor{DoorColor} && Door	&	2.2\%	&	\textbf{48.9\%} &n/a	& 27.5\% & 16.6\% 
\\\arrayrulecolor{white}\hline
\cellcolor{CabinetColor} && Cabinet	&	2.4\%	&	49.8\% &n/a	& 39.8\% & \textbf{52.1\%} 
\\\arrayrulecolor{white}\hline
\cellcolor{BedColor} && Bed	&	2.0\%	&	62.4\% & 52.8\% & \textbf{80.7\%} & 65.9\% 
\\\arrayrulecolor{white}\hline
\cellcolor{DeskColor} && Desk	&	1.7\%	&	36.8\% &n/a	& \textbf{66.7\%} & 58.5\% 
\\\arrayrulecolor{white}\hline
\cellcolor{ToiletColor} && Toilet	&	0.2\%	&	69.9\% &n/a	& 84.8\% & \textbf{86.7\%} 
\\\arrayrulecolor{white}\hline
\cellcolor{SinkColor} && Sink	&	0.2\%	&	39.4\% &n/a	& \textbf{62.8\%} & 53.5\% 
\\\arrayrulecolor{white}\hline
\cellcolor{WindowColor} && Window	&	0.4\%	&	20.1\%	&\textbf{36.1\%} & 23.7\% & 12.5\% 
\\\arrayrulecolor{white}\hline
\cellcolor{PictureColor} && Picture	&	0.2\%	&	\textbf{3.4\%} &n/a	& 0.0\% & 1.8\% 
\\\arrayrulecolor{white}\hline
\cellcolor{BookshelfColor}& & Bookshelf	&	1.6\%	& 	64.6\% &n/a	& \textbf{84.3\%} & 81.0\%
\\\arrayrulecolor{white}\hline
\cellcolor{CurtainColor} && Curtain	&	0.7\%	&	7.0\% &n/a	& \textbf{48.7\%} & 6.1\% 
\\\arrayrulecolor{white}\hline
\cellcolor{ShowerCurtainColor} && Shower Curtain	&	0.04\%	&	46.8\% &n/a  & \textbf{85.0\%} & 48.0\% \\\arrayrulecolor{white}\hline
\cellcolor{CounterColor} && Counter	&	0.6\%	&	32.1\%	&n/a &  \textbf{37.6\%} & 31.6\% 
\\\arrayrulecolor{white}\hline
\cellcolor{RefrigeratorColor} && Refrigerator	&	0.3\%	&	\textbf{66.4\%}	&n/a & 54.7\% & 50.5\% 
\\\arrayrulecolor{white}\hline
\cellcolor{BathtubColor} && Bathtub	&	0.2\%	&	74.3\%	&n/a & \textbf{86.1\%} & 79.1\% \\\arrayrulecolor{white}\hline
\cellcolor{OtherFurnitureColor} && Other Furniture	&	2.9\%	&	19.5\%	&n/a & \textbf{30.7\%} & 30.2\%
\\\arrayrulecolor{black}\hline
&& Weighted Average & & 73.0\% & n/a & \textbf{84.5}\% & 82.6\% \\
&& Unweighted Average & & 50.8\% & n/a & \textbf{60.2\%} & 54.2\% \\
\end{tabular}}}
\end{table}

\subsection{Part Segmentation}
\label{section:EvalPartSegmentation}

We also evaluate our method on a smaller-scale point cloud processing task, model-based semantic part segmentation. To evaluate this, Yi et al. \cite{Yi16} provide a benchmark part segmentation dataset based on ShapeNet. It consists of 50 part categories across 16 types of objects. For example, the car category features part classes: hood, roof, wheel and body. (see Table \ref{table:part_segmentation}).

\subsubsection{Data Preparation}

For this task, we train directly on the provided data without any preprocessing. During training, the input cloud is first rescaled to maximally fit in the unit sphere (2m diameter), then augmented with point dropout and jitter as in the previous task as well as randomly shifted (+/- 5cm) and scaled (+/- 10\%) on the fly.

\begin{figure}[h]
    \centering
  \includegraphics[width=\linewidth]{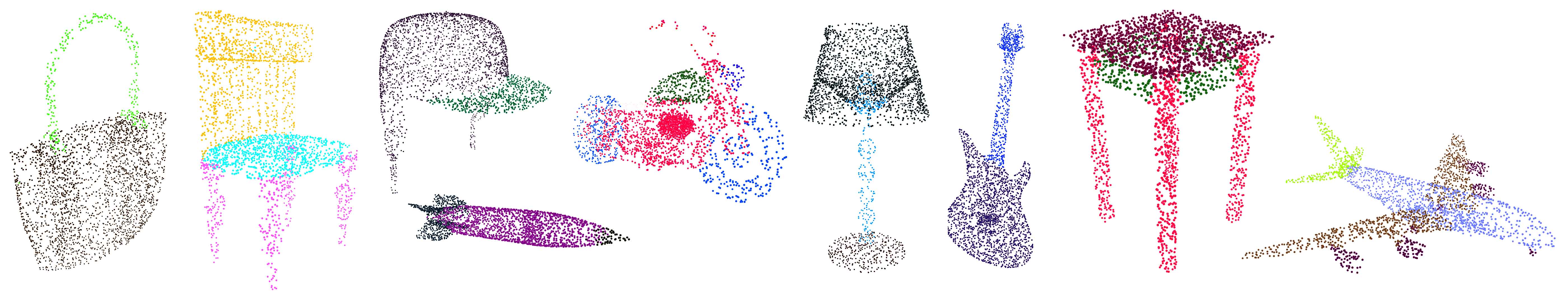} 
    \caption{Qualitative results of the part segmentation test set}
    \label{fig:part_segm_vis} 
\end{figure}

\subsubsection{Network}

The input spatial extent of the network is 2.8m to give every point in the cloud full valid context. The three spatial scales S1, S2, S3 are 10cm, 20cm, 40cm, respectively.  Three $1\times 1\times 1$ layers also follow each abstraction, feature learning and upsampling layer. After upsampling features back to the original sampling resolution, three-nearest-neighbor interpolation is performed in the latent space. Then, like the methods we compare against, the one-hot encoded object class is concatenated to each point's feature vector and followed by three final $1\times 1\times 1$ layers with 50\% dropout between them.

\subsubsection{Results}

The method outperforms the state-of-the-art on this benchmark dataset in 12 out of 16 object categories. Visual examples are given in Figure \ref{fig:part_segm_vis}. 

\begin{table}[h]
\centering
\caption{Results of our FCPN on ShapeNet's part segmentation dataset compared to other state-of-the-art methods. Please note, that we outperform all other methods in 12 out of 16 classes}
\label{table:part_segmentation}
\resizebox{\textwidth}{!}{%
{\renewcommand{\arraystretch}{1.3}%
\begin{tabular}{p{18mm}|c|cccccP{9mm}cccP{9mm}ccccP{8mm}c}
& \textbf{mean} & aero & bag & cap & car & chair & ear\newline phone & guitar & knife & lamp & lap-\newline top & motor & mug & pistol & rocket & skate\newline board & table\\
\hline
Yi \cite{Yangyan2016} & 81.4  & 81.0 & 78.4 & 77.7 & 75.7 & 87.6 & 61.9 & 92.0 & 85.4 & 82.5 & 95.7 & 70.6 & 91.9 & 85.9 & 53.1 & 69.8 & 75.3\\         
SSCNN \cite{Yi2016} &84.7 & 81.6 &81.7 &81.9 &75.2& 90.2& \textbf{74.9} &93.0 &86.1 & \textbf{84.7} &95.6 &66.7 &92.7 &81.6 &60.6 &82.9 &82.1\\
PN \cite{Qi2017} & 83.7 & 83.4 & 78.7 & 82.5 & 74.9 & 89.6 & 73.0 & 91.5 & 85.9 & 80.8 & 95.3 & 65.2 & 93.0 & 81.2 & 57.9 & 72.8 & 80.6\\
PN++ \cite{Qi2017_2} & \textbf{85.1} & 82.4 & 79.0 & \textbf{87.7} & 77.3 & \textbf{90.8} & 71.8 & 91.0 & 85.9 & 83.7 & 95.3 & 71.6 & 94.1 & 81.3 & 58.7 & 76.4 & \textbf{82.6}\\
\hline
Ours & 84.0 & \textbf{84.0} & \textbf{82.8} & 86.4 & \textbf{88.3} & 83.3 & 73.6 & \textbf{93.4} & \textbf{87.4} & 77.4 & \textbf{97.7} & \textbf{81.4} & \textbf{95.8} & \textbf{87.7} & \textbf{68.4} & \textbf{83.6} & 73.4\\
\end{tabular}}}
\end{table}


\subsection{Captioning}
\label{section:EvalCaptioniong}

\begin{figure}[t]
    \centering
 \includegraphics[width=\linewidth]{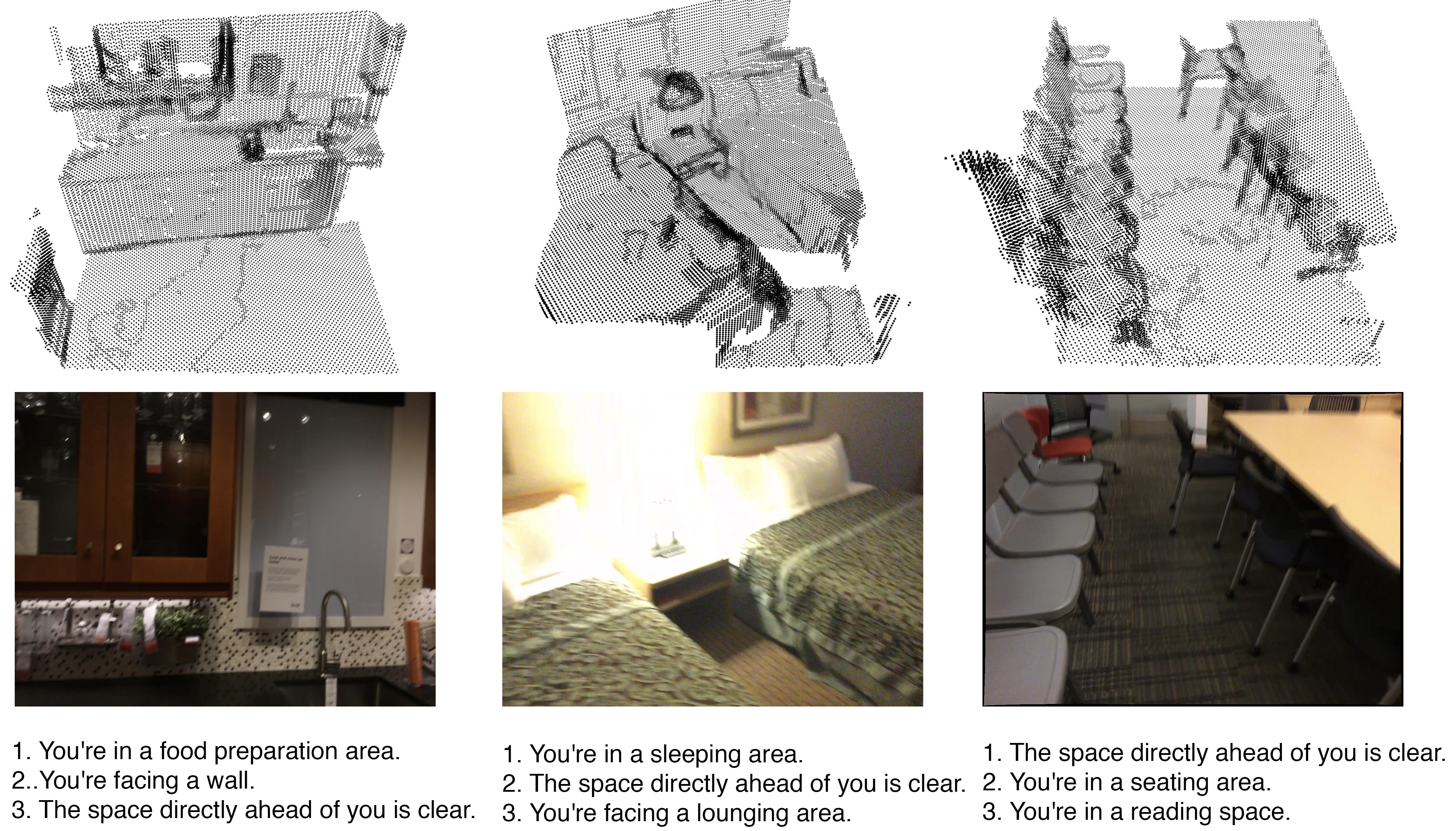} 
    \caption{Example top-3 sentences on 3 frames of the captioning test set. The point cloud in the first row is the input of our captioning model. The RGB image is just illustrated for visualization purposes and not used within our method}
    \label{fig:captiong_result} 
\end{figure}

To demonstrate the usefulness of a spatially ordered output, we evaluate a baseline method for the \textit{3D captioning} task based on the FCPN network. 
To train the captioning model, we take the semantic voxel labeling network and replace the final upsampling layer and subsequent convolutional layers with three fully-connected layers. We freeze the weights of the semantic voxel labeling network and train only the fully-connected layers on this task. Once again, captions are weighted by the inverse log of their frequency in the training set. We consider the top-3 most confident captions produced by the network. Examples are shown in Figure \ref{fig:captiong_result}. Differently from standard image-based captioning, the provided results hint at how the 3D captioning output together with the proposed network can usefully summarize the relevant scene geometry with respect to a specific viewpoint to aid navigation and interaction tasks. Additional results are provided in the supplementary material. 

\subsection{Memory and Runtime}

To complement our evaluation, we estimate the memory footprint and inference speed of our method as a function of the input cloud size (both spatially and in terms of point count), by processing clouds of increasing point counts and surface area on a Titan Xp GPU. The results in Table \ref{table:runtime} validate our claim that the proposed method can efficiently process large-scale point clouds, with clouds 5x as large with 10x the amount of points requiring only 40\% more memory.

\begin{table}[h]
\setlength{\tabcolsep}{1em}
\centering
\caption{Memory Consumption Evaluation}
\label{table:runtime}
{%
{\renewcommand{\arraystretch}{1.1}%
\begin{tabular}{l l l l}
\textbf{Point Count} & \textbf{Surface Area} & \textbf{Forward Pass} & \textbf{Memory} \\
\hline
150k & $80m^2$ & 9.1s & 9033 MB\\
36k & $36m^2$ & 2.9s & 8515 MB\\
15k & $16m^2$ & 0.57s & 6481 MB\\                 
\end{tabular}}}
\end{table}

\section{Conclusions}

In this paper we presented the first fully-convolutional neural network that operates on unordered point sets. We showed that, being fully convolutional, we are able to process by far larger spaces than the current state of the art in a single shot. We further show that it can be used as a general-purpose feature descriptor by evaluating it on challenging benchmarks at different scales, namely semantic scene and part-based object segmentation. This shows its proficiency in different tasks and application areas. Further, since it learns a spatially ordered descriptor, it opens the door to higher level scene understanding tasks such as captioning. As for future work, we are interested in exploring a wider range of semantic classes and using our network to train a more expressive language model.

\bibliographystyle{splncs}
\bibliography{egbib}

\end{document}